\documentclass[11pt]{article}

\usepackage[final]{acl}

\usepackage{newtxtext}
\usepackage{newtxmath}
\usepackage[T1]{fontenc}
\usepackage[utf8]{inputenc}
\usepackage{amsmath}

\usepackage{amssymb}

\usepackage{latexsym}
\usepackage{booktabs}
\usepackage{multirow}
\usepackage{arydshln}
\usepackage{enumitem}
\usepackage{placeins}

\usepackage{algorithm}
\usepackage{algpseudocode}

\usepackage{microtype}
\usepackage{inconsolata}

\usepackage{twemojis}
\usepackage{worldflags}

\usepackage{graphicx}

\title{\scalebox{2}{\twemoji{banana}} Is Sentiment Banana-Shaped? Exploring the Geometry and Portability of Sentiment Concept Vectors}
    
\author{
  \textbf{Laurits Lyngbaek\textsuperscript{*}},
  \textbf{Pascale Feldkamp\textsuperscript{*}},
  \textbf{Yuri Bizzoni\textsuperscript{}},
  \\
  \textbf{Kristoffer L. Nielbo\textsuperscript{}},
  \textbf{Kenneth Enevoldsen\textsuperscript{}}.
  \\
  Aarhus University, Aarhus, Denmark
  \\
  \textsuperscript{*}Shared First Authorship
}

\begin{document}
\maketitle
\begin{abstract}
% importance
%Sentiment analysis in humanities research %literary and historical texts 
%often requires continuous scores.
Use cases of sentiment analysis in the humanities often require contextualized, continuous scores.
% motivation
Concept Vector Projections (CVP) offer a recent solution: by modeling sentiment as a direction in embedding space, they produce continuous, multilingual scores that align closely with human judgments. Yet the method’s portability across domains and underlying assumptions remain underexplored.
% what we do
We evaluate CVP across genres, historical periods, languages, and affective dimensions, finding that concept vectors trained on one corpus transfer well to others with minimal performance loss. 
To understand the patterns of generalization, we further examine the linearity assumption underlying CVP. 
Our findings suggest that while CVP is a portable approach that effectively captures generalizable patterns, 
its linearity assumption is approximate, pointing to potential for further development. Code available at: github.com/lauritswl/representation-transfer
\end{abstract}

%Extending the method beyond valence, we find comparable—though weaker—generalization for arousal and dominance.

\section{Introduction and Related Works}

Sentiment Analysis approaches to data in the Humanities %-- especially literary data -- 
often need continuous sentiment scores to develop meaningful models of texts, for tasks such as tracing the ``sentiment arc'' of a story \cite{jockers_novel_2014, reagan_emotional_2016, cellier_prediction_2016, bizzoni_sentimental_2023}, gauging sentiment fluctuations in news \cite{daudert_exploiting_2021} or modeling changes in online discourse \cite{xie_tracking_2025}, but existing tools struggle to capture the necessary nuances effectively. %in texts or to contextualize single words' continuous sentiments to sentences and paragraphs. 
Many dictionary-based methods are continuous, but struggle with extended context, whereas Transformer models produce binary or trinary outputs that only approximate continuous sentiments through post-hoc adjustments \cite{bizzoni_comparing_2023, lyngbaek_continuous_2025}.

\begin{figure*}
    \centering
    \includegraphics[width=0.791\linewidth]{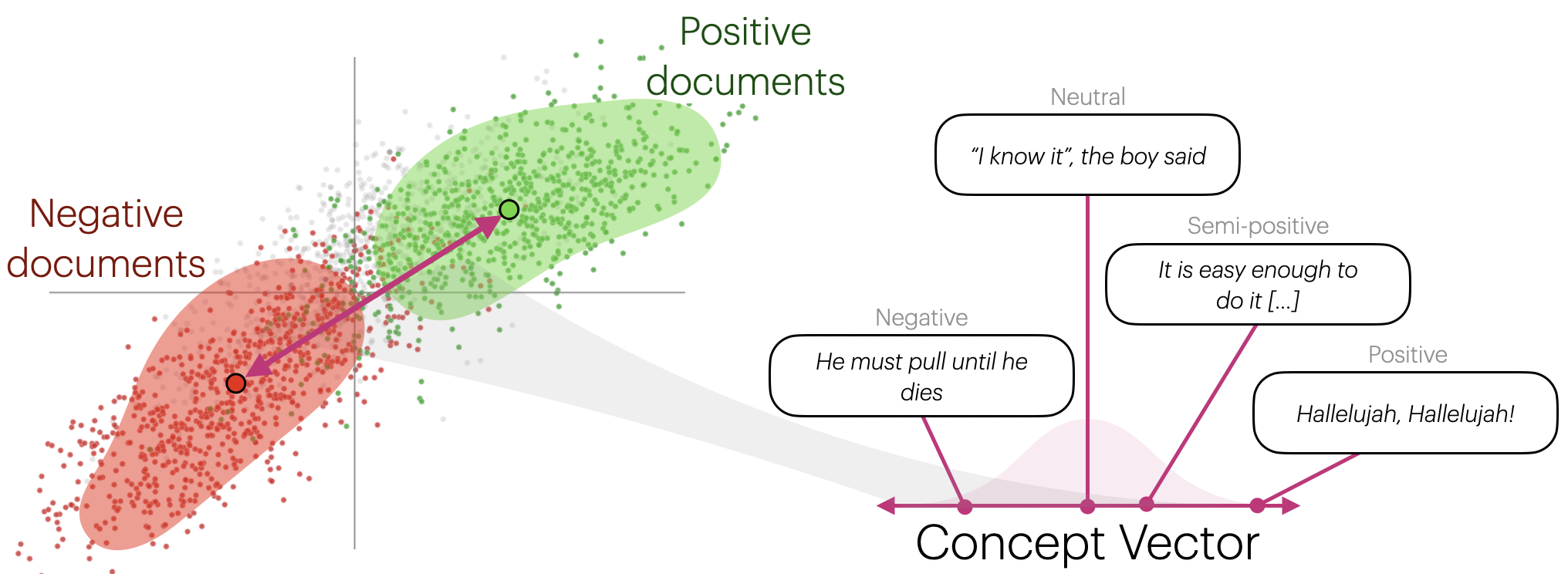}
    \caption{A visualization of how the Concept Vector Projection is constructed. It shows how to construct a positive-negative concept vector to predict sentiment in an unlabeled corpus in a continuous space.}
    \label{fig:concept-vector-intro}
\end{figure*}

A recent alternative \cite{lyngbaek_continuous_2025} uses a projection-based method in a homogeneous semantic space to generate continuous sentiment scores that align with human judgments and match or surpass Transformer-based methods on literary data, while producing smoother distributions. This approach, called Concept Vector Projection (CVP), rests on the ``linear representation hypothesis'' \cite{park_linear_2024}: the idea that semantic concepts, such as sentiments, can be represented linearly in embedding space \cite{wehner_taxonomy_2025, vu-parker-2016-k, li-etal-2021-implicit, zhao2024beyond}. Under this idea, a given semantic concept corresponds to a \textit{direction} in the embedding space, so that moving further along this direction increases its intensity (see \autoref{fig:concept-vector-intro}).% %Projecting sentences onto this vector yields a continuous measure of valence (Figure \ref{fig:concept-vector-intro}).

While studies have validated this idea at various levels of abstraction \cite{lyngbaek_continuous_2025, wehner_taxonomy_2025, vu-parker-2016-k, li-etal-2021-implicit, zhao2024beyond}, its \textit{portability} across different data domains and semantic dimensions remains underexplored. 
Literary texts, blogs, newspapers, and social media differ in style and affective expression \cite{feldkamp2024sentiment, vishnubhotla-etal-2024-emotion},\footnote{How much domains differ varies. For example, if using a model fine-tuned on Twitter posts, poetry shows the weakest correlation with human ratings, prose falls in the middle, and Facebook posts show the strongest correlation \cite{feldkamp2024sentiment}.} and language or period differences can complicate the transfer.
The general trend in sentiment analysis has been to assume non-portability and train or fine-tune specialized models for specific domains, languages and historical variants \cite{allaith_sentiment_2023, schmidt_evaluation_2018} -- models that are then difficult to adapt for other use cases. %Moreover, such models still entails the issue of non-continuous distributions.
%A model's ability to generalize suggests that its projection captures deeper, domain-independent properties of sentiment.

In this work, we test the CVP's portability across three datasets, spanning genres (social media to letters), periods (1798-2013), and languages (English and Danish). %(Table \ref{tab:dataset_stats}). 
After presenting the Data (\autoref{sec:data}) and Methods (\autoref{sec:methods}), we test the CVP through several cross-dataset experiments (\autoref{sec:portability}) to assess whether the resulting scores retain their alignment with human judgments.
We also explore the portability of the CVP beyond valence -- to related affective dimensions, such as arousal and dominance (\autoref{sec:beyond_valence}), and consider whether imperfect linearity in the projections might be the cause of some of the method's inaccuracies (\autoref{sec:linearity}).

\section{Datasets}
\label{sec:data}

To represent diversity across the literary and non-literary domains, we select three datasets that span different genres, domains, languages, and historical periods, all using a human-annotated continuous scale.

\paragraph{}
\textbf{Emobank} \cite{buechel-hahn-2017-emobank} contains sentences from the MASC dataset annotated according to the Valence-Arousal-Dominance (VAD) scheme \cite{mehrabian1974approach}. The dataset includes: Letters, Blog, Newspaper, Essays, Fiction, and Travel guides.

% facebook
\paragraph{}
\textbf{Facebook} \cite{preotiuc-pietro-etal-2016-modelling} consists of status updates collected by \cite{kosinski_private_2013} and annotated for valence and arousal.

\paragraph{}
\textbf{Fiction4} \cite{feldkamp2024sentiment} comprises literary texts spanning four genres and two languages (English/Danish) from the 19\textsuperscript{th} and 20\textsuperscript{th} centuries.
It consists of three main authors -- Sylvia Plath (poetry), Ernest Hemingway (prose), and H.C. Andersen (fairytales) -- and hymns from Danish official church hymnbooks (published 1798-1873). Two or more human annotators scored each sentence (/line, for poetry) for valence.\footnote{Although lower than Facebook posts, IRR for Fiction4 ($\alpha$=0.67) is high for continuous annotations of literary texts. Humans rarely reach $\alpha>0.80$ for polarity tagging on \textit{non-literary} texts \cite{wilson_recognizing_2005} and achieve lower IRR for continuous scales on literary texts \cite{batanovic_versatile_2020, rebora_comparing_2023}.}

\begin{table}[htbp]
\centering
\resizebox{\columnwidth}{!}{
\begin{tabular}{lcrc}
\toprule
\textbf{Dataset} & \textbf{Period} & \textbf{Sentences} & \textbf{Kripp. $\alpha$} (Scale) \\
\midrule
\textbf{EmoBank} & 1990--2008 & 10,062 &  \\
\hspace{1.5em} Valence &  &  & .34 (1-5) \\
\hspace{1.5em} Arousal &  &  & .25 (1-5) \\
\hspace{1.5em} Dominance &  &  & .22 (1-5) \\
\textbf{Facebook} & 2012--2013 & 2,895 & \\
\hspace{1.5em} Valence &  &  & .72 (1-9) \\
\hspace{1.5em} Arousal &  &  & .82 (1-9) \\
\textbf{Fiction4} & 1798--1965 & 6,300  \\
\hspace{1.5em} Valence &  &   & .67 (0-10) \\
\bottomrule
\end{tabular}
}
\caption{Summary of annotated corpora. We report sentence counts, average length, and inter-rater agreement ($\alpha$). The total number of sentences considered is $n=19,257$. Full breakdown of subgenres (in Fiction4 and Emobank) and number of annotators in \autoref{sec:data-details}.
}
\label{tab:dataset_summary}
\end{table}

\section{Methods}
\label{sec:methods}

To construct the concept vector, we follow the approach introduced by \cite{lyngbaek_continuous_2025}, where a pre-trained sentence-embedding model $\mathbf{M}$ embeds a set of source$_{negative}$ and target$_{positive}$ exemplar sentences. We compute the mean embeddings of source$_{negative}$ and target$_{positive}$ examples and define the concept vector $\hat{\mathbf{v}}$ as the unit vector of the difference between mean embeddings. The assumption is that this averaging will reduce non-sentiment information to Gaussian noise with a mean of zero, leaving the sentimental signal behind \cite{kim2018interpretability, zhao2024beyond}.
With this method, we score a sentence $s$ by projecting its embedding onto the concept vector $\hat{\mathbf{v}}$ via the dot product $\mathbf{M}(s) \cdot \hat{\mathbf{v}}$, yielding a continuous sentiment score. We normalize the scores using a z-score normalization. We define the details for the CVP algorithm in \autoref{sec:algorithm}.\footnote{Implementation available at \href{https://github.com/lauritswl/representation-transfer}{representation-transfer}}
To define source$_{negative}$ and target$_{positive}$, we set sentiment thresholds relative to each corpus’ valence distribution. Sentences at least one standard deviation above the mean are positive, sentences below by one standard deviation are negative, and the rest are neutral (for the formalization, see \autoref{sec:polarity-procedure}). We estimate \textit{concept vectors} from these positive–negative contrasts, capturing the extremes rather than absolute ratings. This approach yields comparable sentiment contrasts across datasets with different scales and distributions. For testing linearity, we created three concept vectors: positive–negative, negative–neutral, and neutral–positive.

\subsection{Model}
To allow for comparability with previous works \cite{lyngbaek_continuous_2025}, we use the embedding model paraphrase-multilingual-mpnet\footnote{\href{https://huggingface.co/sentence-transformers/paraphrase-multilingual-mpnet-base-v2}{sentence-transformers/paraphrase-multilingual-mpnet-base-v2}} \cite{reimers-2019-sentence-bert}, a 278M-parameter model based on a mean-pooled BERT architecture optimized for sentence similarity via Siamese and Triplet networks. This model is notable for its multilingual capabilities, previous performance \cite{lyngbaek_continuous_2025}, and excellent size-to-performance ratio.\footnote{A larger model may increase model correlation with human scores at the expense of computation budget.}

% Sentences whose valence lies at least one standard deviation above the mean are treated as positive, those at least one standard deviation below the mean as negative, and the remaining sentences as neutral (see \autoref{sec:polarity-procedure}). Concept vectors are then estimated from the resulting positive–negative contrasts. In other words, the concept vectors are learned from the extremes of each corpus’ valence distribution, rather than from absolute rating values.
% This procedure yields comparable sentiment contrasts across datasets with different annotation scales, distributions, and historical or generic profiles. 
% For our first step, we created 3 concept vectors: for positive–negative, negative-neutral, and neutral-positive contrasts.

\section{Results}

\subsection{Portability}
\label{sec:portability}

Our results show that the projection method is robust: continuous valence scores remain well-aligned to human scores across all three datasets and their constituent subgenres (\autoref{tab:subcategory-correlations}), suggesting that the approach captures generalizable sentiment patterns beyond the idiosyncrasies of literary, journalistic, or social media language. It highlights the portability of continuous sentiment scoring across genres, which can be crucial for research spanning multiple text types or for investigating historical and contemporary corpora side by side.

\begin{table}[ht]
\centering
%\resizebox{\columnwidth}{!}{
\scriptsize
\begin{tabular}{lccc}
\toprule
\textbf{Dataset} & \multicolumn{3}{c}{Correlation, when trained on:} \\
\cmidrule(lr){2-4} & \textbf{Fiction4} & \textbf{Emobank} & \textbf{Facebook} \\
\midrule
Fiction4  & \textbf{0.66} & \underline{0.65} &  0.64\\
Emobank   & \underline{0.67} & \textbf{0.70} & 0.66 \\
Facebook  & \underline{0.66} & \underline{0.66} & \textbf{0.68} \\
\bottomrule
\end{tabular}
%}
\caption{Spearman correlations between human and projected valence scores across corpora. Values indicate correlations when trained on the indicated corpus (columns) and tested on itself or another corpus (rows).}
\label{tab:subcategory-correlations}
\end{table}

\begin{table}[h]
    \centering
    \scriptsize
    \begin{tabular}{lrrr}
    \toprule
         &  \textbf{Emobank} & \textbf{Facebook} & \textbf{Fiction4} \\
        \midrule
        Valence &.71$\pm$.02(.70)&.70$\pm$.02(.68)&.66$\pm$.02(.66)\\
        Arousal &.36$\pm$.02(.42)&.65$\pm$.02(.67)& \\
        Dominance&.35$\pm$.01(.37)& & \\
        \bottomrule
    \end{tabular}
    \caption{Cross-validation of Spearman correlations between CVP scores and human scores for valence, arousal, and dominance per corpus. The scores are the mean correlation obtained from a five fold analysis, with a standard deviation notated by $\pm$. The score parenthesis indicates the Spearman correlation obtained when no split was conducted. Only Emobank has human scores of all V-A-D labels.}
    \label{tab:VAD}
\end{table}

\subsection{Beyond Valence}
\label{sec:beyond_valence}
To test the CVP's ability to generalize beyond valence -- which refers to the positivity/negativity spectrum -- we tested the approach on semantic properties associated with valence in sentiment analysis: arousal and dominance. Arousal refers to the intensity of the concept conveyed by a given word (\textit{ecstatic} and \textit{serene} are both positive, but the first word elicits a higher arousal); dominance refers to the amount of control associated with a term (\textit{angry} and \textit{helpless} are both negative, but the first word has more dominance). We find that CVP generalizes well for these subtler concepts (Table \ref{tab:VAD}) with similarly continuous distributions (see \autoref{sec:arousal_dominance_vizes}), without reaching the performance achieved on valence.

\subsection{Linearity assumption}
\label{sec:linearity}
\begin{figure}
    \centering
    \includegraphics[width=1\linewidth]{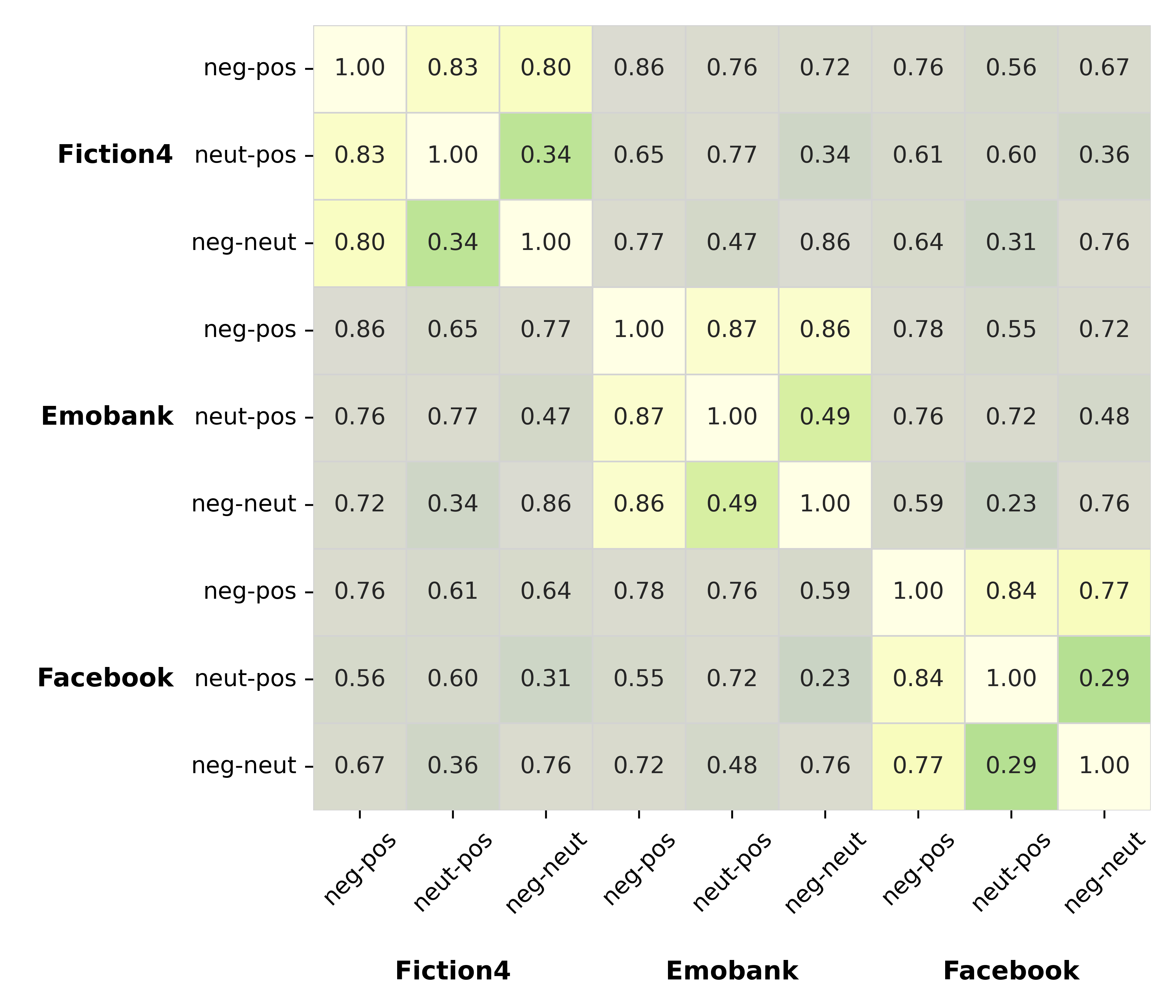}
    \caption{Cosine similarity between Concept Vectors for each corpus (values in each cell). Internal correlations among neg-pos, neut-pos, and neg-neut pairs are strong, with neut-pos and neg-neut closer to neg-pos, reflecting a centrality of the negative–positive axis across corpora.}
    \label{fig:cosine-matrix}
\end{figure}

CVP treats sentiment as linear in embedding space: negative and positive extremes form the main axis, with neutral sentences in the middle. We create similar vectors with negative–neutral and neutral–positive extremes, and examine the cosine similarity between all three vectors (\autoref{fig:cosine-matrix}). Across corpora, the negative–positive axis aligns most strongly with the other two axes, consistent with a geometrical interpretation that neutral texts are located close to the axis, but vary along an undefined semantic dimension, so that the three vectors will form a triangle outlining the centroids of the three classes. %, if the neutral centroid isn't located directly on the axis.
We explore the geometry of Fiction4's valence space by creating a two-dimensional basis that visualizes the data. We define our first dimension as the negative-positive vector, $\mathbf{v}_{np}$. The second semantic dimension we define as the neutral-component, $\mathbf{v}_{nc}$. The neutral-component vector captures the remaining semantic information encoded in the neutral centroid. 

%% INCORECT definition from preprint
%By projecting the neutral centroid $C_{neu}$ onto the negative-positive vector, we get a scalar $k$ of the negative-positive vector $\mathbf{v_{np}}$. Using the negative centroid $C_{neg}$ and the scaled negative-positive vector $k \cdot \mathbf{v_{np}}$, we can define the projection of the neutral centroid onto the negative-positive vector as: $proj_{\vec{v_{np}}}(C_{neu})=C_{neg}+k \cdot \mathbf{v_{np}}$. We use this projected point to define the neutral-component vector $\mathbf{v_{nc}}$ as the vector going from the neutral centroid $C_{neu}$ to $proj_{\vec{v_{np}}}(C_{neu})$, as: 
%$v_{nc} = C_{neu}-proj_{\vec{v_{np}}}(C_{neu})$

To define the neutral component, we treat the problem geometrically in an affine space. Because an affine space has no natural origin, projections cannot be applied directly to centroids. Instead, we work with \emph{difference vectors}, which encode relative positions between centroids. Let $\mathbf{v}_{np}$ denote the vector from the negative to the positive centroid and define the corresponding unit direction
\[
\hat{\mathbf{v}}_{np} = \frac{\mathbf{v}_{np}}{\|\mathbf{v}_{np}\|}.
\]
We also define the vector from the negative to the neutral centroid as $\mathbf{v}_{nn}$. Projecting $\mathbf{v}_{nn}$ onto $\hat{\mathbf{v}}_{np}$ gives the scalar projection
\[
k = \mathbf{v}_{nn} \cdot \hat{\mathbf{v}}_{np}.
\]
This scalar specifies the relative position of the neutral centroid along the negative--positive axis, yielding the projected component
\[
k \, \hat{\mathbf{v}}_{np}.
\]

Finally, we construct the \emph{neutral component vector} by removing this projected information:
\[
\mathbf{v}_{nc} = \mathbf{v}_{nn} - k \, \hat{\mathbf{v}}_{np}.
\]
This residual vector represents the component of the neutral direction that is orthogonal to the sentiment axis.
\begin{figure}
    \centering
    \includegraphics[width=0.95\linewidth]{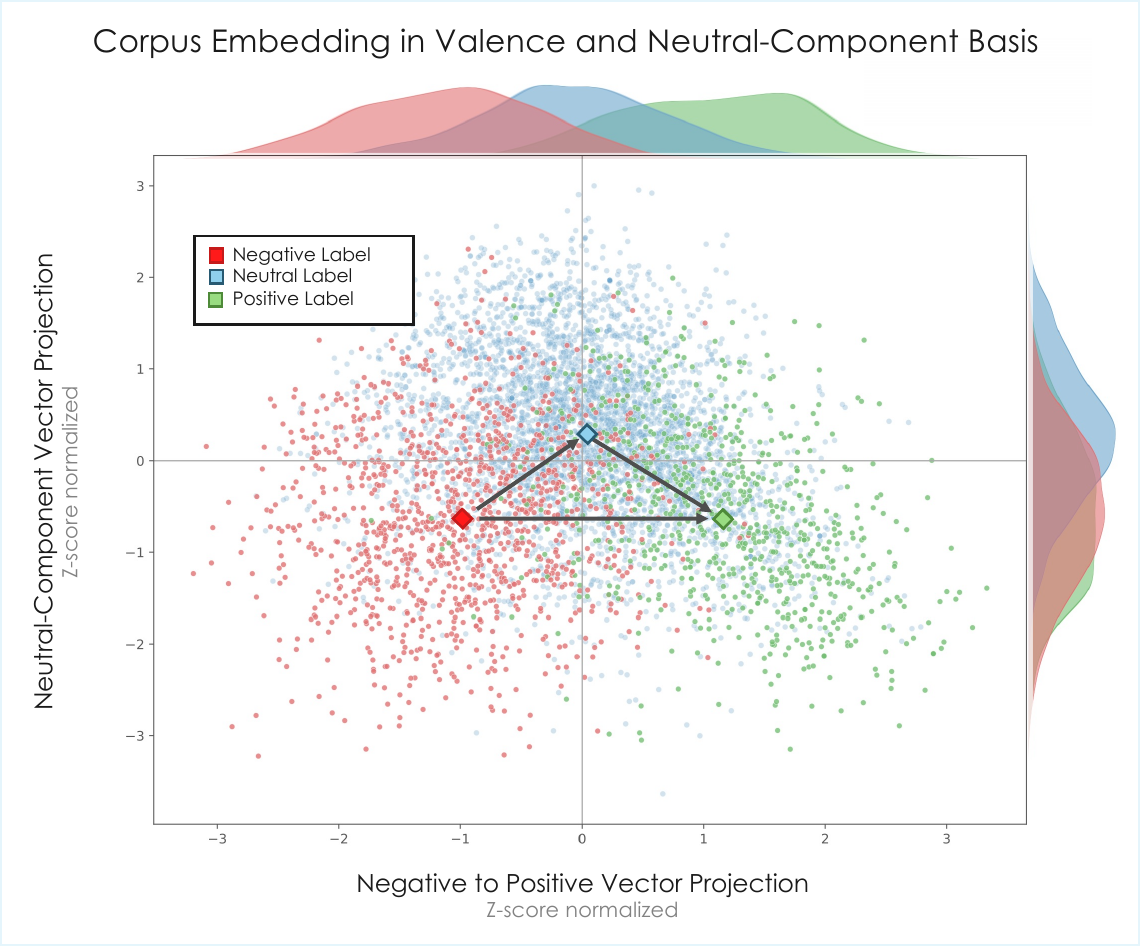}
    \caption{Scatterplot visualizing the Fiction4 embeddings projected onto the Fiction4 Pos-Neg Sentiment vector and the corresponding Neutral-Component. The Margin plots are Kernel Density Estimations of the label distributions. All dimensions are Z-score normalized to make the projections interpretable.}
    \label{fig:banana}
\end{figure}
We use this constructed basis to visualize the geometric structure of sentiment embeddings, as seen in \autoref{fig:banana}. This result aligns with the high-dimensional cosine-similarity observed in \autoref{fig:cosine-matrix}. We see that our embeddings tend to be linear, but that neutral embeddings encode spurious information that remains unaccounted for in the sentiment direction. This property gives the centroids a triangular shape -- and structures the Fiction4 embeddings as a banana-shaped manifold.

\section{Discussion \& conclusions}

% For an estimate of how other models do on Emobank and Fiction4, see \cite{lyngbaek_continuous_2025}.

We find that Concept Vector Projections transfer well across genres, periods, and languages — a vector derived from a corpus including 19\textsuperscript{th}-century Danish hymns predicts sentiment in contemporary Facebook posts nearly as well as one trained on in-domain data. This portability suggests CVP captures generalizable properties of how sentiment is encoded in embedding space, rather than domain-specific patterns. The approach also extends to arousal and dominance, though with reduced performance, consistent with valence being the most reliable and consistent of the three affective dimensions \cite{warriner_norms_2013}. For researchers working with historical or low-resource corpora, this means domain-specific training data may not be necessary -- as suggested in \citet{lyngbaek_continuous_2025}.

Our geometric analysis shows that the linearity assumption is approximate: neutral sentences do not lie exactly on the positive–negative axis but form a continuous, banana-shaped curve. This suggests neutrality encodes semantic content beyond the absence of valence -- a property that future methods might exploit.

\section*{Limitations}

While the goal of this work, is not to explore how to optimize the performance of the CVP, but examine its implications, we only examine one model to ensure comparability with \citet{lyngbaek_continuous_2025}. Further analysis should explore alternative models as indicate evidence suggests that newer models like EmbeddingGemma \cite{vera2025embeddinggemmapowerfullightweighttext} might surpass the one currently used.

To examine the CVP ability to generalize to the related concepts arousal and dominance, we utilize the dataset itself as the source dataset for deriving the concept vector. This likely leads to a modest overestimation of the correlation as we see for valence in \autoref{tab:subcategory-correlations} and arousal in \autoref{app:portability-arousal}. 

Our cross-lingual evaluation, while leveraging a multilingual embedding model, is restricted to Danish and English. These languages, though differing in resource availability, belong to the same language family; generalization to typologically distinct languages remains untested.

    \section*{Acknowledgments}

This work was partially supported by the Danish National Research Foundation (DNRF193) through TEXT: Center for Contemporary Cultures of Text, Aarhus University. 
% UCloud
% Anything else?

\newpage
\bibliography{custom}

\newpage
\onecolumn
\appendix

\clearpage

\section{Vector polarity selection procedure}
\label{sec:polarity-procedure}

For each text unit \(i\) in a corpus, we compute its valence score \(v_i\). Let \(\mu\) and \(\sigma\) denote the mean and standard deviation of valence scores across the corpus. Polarity labels are assigned as follows:

\small
\[
\text{label}_i =
\begin{cases}
\text{positive}, & \text{if } v_i \ge \mu + \sigma\\
\text{negative}, & \text{if } v_i \le \mu - \sigma\\
\text{neutral}, & \text{otherwise}
\end{cases}
\]
\normalsize

This scheme assigns a label based on deviation from the corpus mean by one standard deviation.

\section{CVP algorithm}
\label{sec:algorithm}

The following algorithm formally describes the procedure for defining and applying a concept vector by using labeled sentence embeddings. 

\begin{algorithm}
    \small
  \caption{Concept Vector Projection}\label{conceptvector}
  \begin{algorithmic}[0]
  %-------------- Input & Output ----t-------------
  % \State \textbf{Input:} \\
  % $\mathcal{M}$ = Language Model \\
  % $\mathcal{S}$ = A set of categorically labeled sentences  $s_i \in \{\text{positive}^+, \text{negative}^-, \text{neutral}^\varnothing, \text{unknown}^?\}$
  \State \textbf{Input:} \\
$\mathbf{M}$ = Language Model \\
$\mathbf{S}$ = A set of sentences $s_i$, labeled via mean ± SD thresholds for valence:
$s_i \in \{\text{positive}^+, \text{negative}^-, \text{neutral}^\varnothing, \text{unknown}^?\}$
    % i.e. $s_i \in \{s_i^{c^{+}} \cup s_i^{c^{-}} \cup, s_i^{c^{\varnothing}} \cup  s_i^{c^{?}} \}$
  \State \textbf{Output:} \\
  $\hat{\mathbf{v}}$ = Concept vector  \\
  score($s_i$) = projection scores for unknown sentences
 \State \textbf{Computation:}
 \end{algorithmic}
  %-------------- Computation --------------------
  \begin{algorithmic}[1]
  \State Embed all sentences: $\mathbf{e}_i = \mathbf{M}(s_i)$
  \State $P^+ \gets \{\mathbf{e}_i \mid s_i = \text{positive}\}$
  \State $N^- \gets \{\mathbf{e}_i \mid s_i = \text{negative}\}$
  \State Compute means: $\mathbf{\mu}_{s^{+}} = \text{mean}(P^+)$, $\mathbb{\mu}_{s^{-}} = \text{mean}(N^-)$
  \State Compute concept vector: $\vec{\mathbf{v}} = \vec{\mu_{s^{+}}} - \vec{\mu_{s^{-}}}$
  \State Normalize: $\hat{\mathbf{v}} = \frac{{\mathbf{v}}}{\|{\mathbf{v}}\|}$
  %-------------- Projection ----------------------
  \For{\textbf{each} embedding $\mathbf{e}_i$} 
    \State $\text{score}(s_i) = \mathbf{e}_i \cdot \hat{\mathbf{v}}$   
  \EndFor
  \State Standardize scores: $\frac{\text{score}(s_i)-mean(\text{score}(s_i))}{std(\text{score}(s_i))}$\hfill // Embedding projection
  \end{algorithmic}
\end{algorithm}

\section{Performance baseline}
\label{sec:baseline_correlations}

To contextualize the correlations between the CVP and human scores, we also include the correlations between a transformer-based model and human scores. We choose the best-performing model in \citet{lyngbaek_continuous_2025}, the multilingual \texttt{cardiffnlp/xlm-roberta-base-sentiment-multilingual} (here abbreviated xlm-R-b)\footnote{\url{https://huggingface.co/cardiffnlp/xlm-roberta-base-sentiment-multilingual}}, which is an xlm-roberta model finetuned for sentiment on Twitter data \cite{barbieri_xlm-t_2022}. The model's output was transformed using its confidence scores, consistent with the approach in \citet{lyngbaek_continuous_2025} and \citet{bizzoni_comparing_2023}. We do not compare to continuous dictionary-based sentiments approaches like VADER, but for a comparison against these methods, we refer to \citet{lyngbaek_continuous_2025}.

\begin{table*}[ht]
\centering
\small
\begin{tabular}{ll|c|ccc}
\toprule
\textbf{Dataset} & \textbf{Subcategory} & \textbf{xlm-R-b} & \multicolumn{3}{c}{\textbf{Correlation, when trained on:}} \\
\cmidrule(llr){4-6} & & & \textbf{Fiction4} & \textbf{Emobank} & \textbf{Facebook} \\
\midrule
\textbf{Fiction4} & \textit{overall} & 0.60 & \textbf{0.66} & \underline{0.65} &  0.64\\
         & fairytales & 0.62 & \textbf{0.67} & \underline{0.64} & 0.61 \\
         & hymns & 0.59 &  \textbf{0.67} & \underline{0.66} & 0.62 \\
         & poetry & 0.57 & \textbf{0.72} & \underline{0.71} & 0.68 \\
         & prose  & 0.61 & \textbf{0.64} & 0.62 & \underline{0.63} \\
\midrule
\textbf{Emob.}  & \textit{overall} & 0.65 & \underline{0.67} & \textbf{0.70} & 0.66 \\
         & SemEval    &   0.64 & \underline{0.66} & \textbf{0.71} & 0.65 \\
         & blog       &   0.65 & 0.64 & \underline{0.68} & \textbf{0.69} \\
         & essays     &  0.58 & \underline{0.59} & \textbf{0.63} & 0.55 \\
         & fiction    &  0.56 & \underline{0.67} & \textbf{0.69} & \underline{0.67} \\
         & letters    &   \underline{0.68} & \underline{0.68} & \textbf{0.71} & 0.66 \\
         & newspaper   & 0.65 & \underline{0.67} & \textbf{0.69} & 0.65 \\
         & travel-guides  & 0.49 & 0.56 & \underline{0.58} & \textbf{0.59} \\
\midrule
\textbf{FB} & \textit{overall} & 0.74 & \underline{0.66} & \underline{0.66} & \textbf{0.68} \\
\bottomrule
\end{tabular}
\caption{Correlations with human and projected valence scores across corpora. Values indicate correlations when trained on the indicated corpus (columns) and tested on the datasets overall and across subgenres (rows). Correlation of the transformer-based model and human score is indicated in column \textbf{xlm-R-b}.}
\label{tab:subcategory-correlations-detailed}
\end{table*}

\begin{figure}[h]
    \centering
    \includegraphics[width=0.5\linewidth]{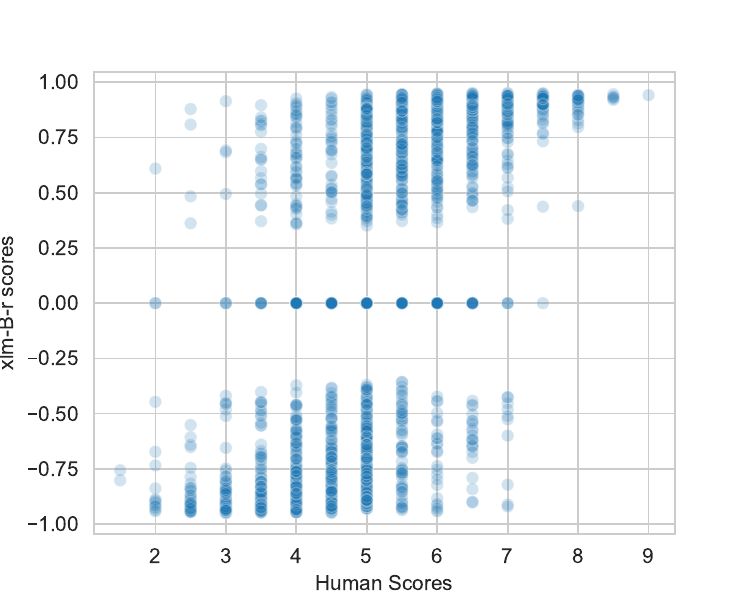}
    \caption{Correlation (Spearman's $\rho$) between transformer model (xlm-R-b) and human scores in the Facebook dataset.}
    \label{fig:corr_facebookxtransformer}
\end{figure}

\begin{figure}
    \centering
    \includegraphics[width=0.9\linewidth]{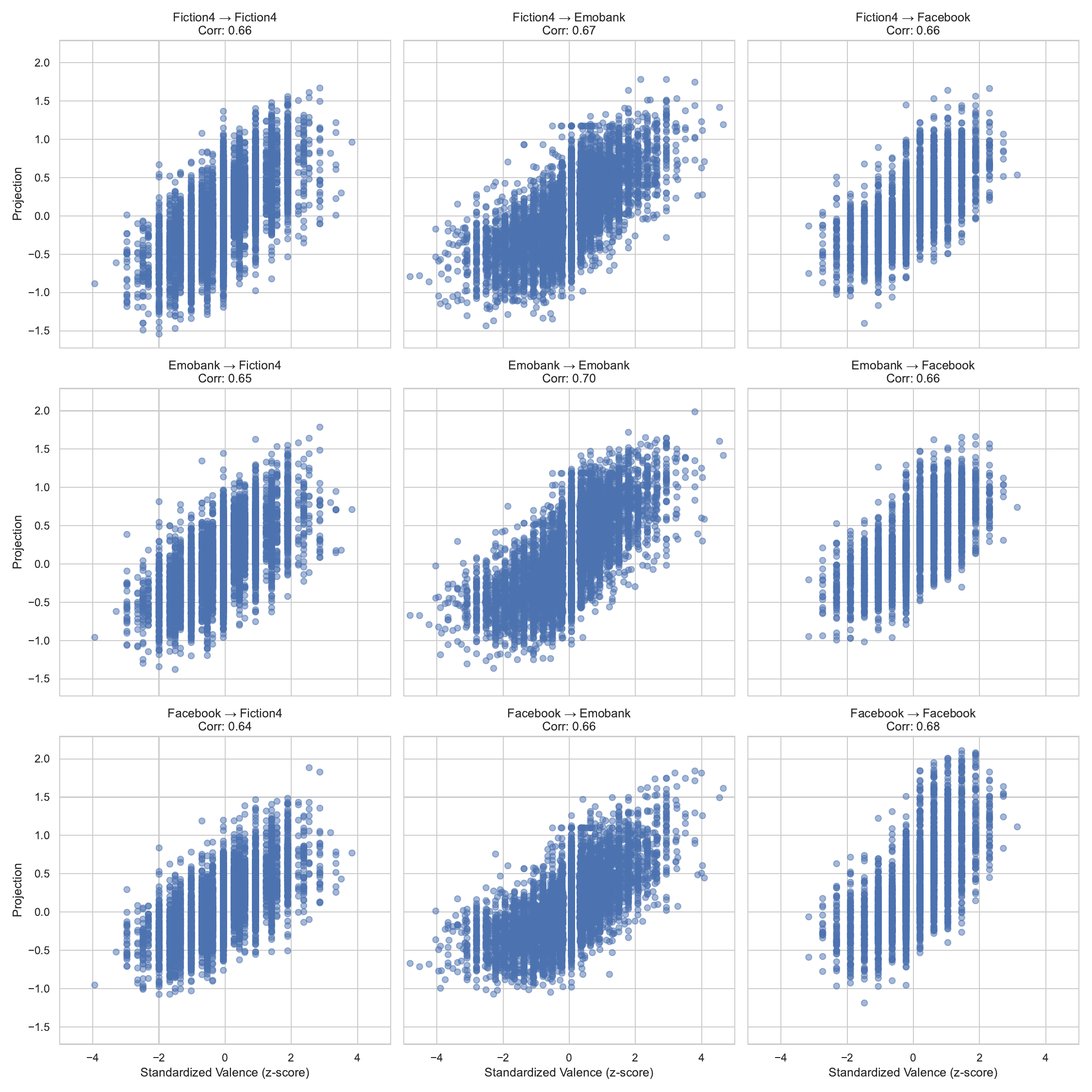}
    \caption{Relation between Concept Vector Projection scores (y-axis) and human scores (x-axis) on standardized valence across datasets. On top of each figure, the training set (on the left of the arrow) and the test set (on the right of the arrow) are shown.%. For each vertical column, you see the method tested on a dataset when trained on each dataset. For example, for the left-hand column (from top to bottom): tested on Fiction4, trained on itself, trained on Emobank, trained on Facebook.
    }
    \label{fig:scatters}
\end{figure}

Note that while the \textbf{xlm-R-b} model performs better than the Concept Vector Projection on Facebook data in terms of Spearman's $\rho$ (see \autoref{tab:subcategory-correlations}), the distributions of these scores remain pseudo-trinary (see the \autoref{fig:corr_facebookxtransformer}), unlike the distribution of the Concept Vector Projection's scores (\autoref{fig:scatters}).

\FloatBarrier
\section{Data details}
\label{sec:data-details}

A note on \textit{EmoBank} categories: \textit{Essays} include, i.e., ``A Brief History of Steel in Northeastern Ohio''. \textit{Fiction} comprises prose pieces, i.e., Richard Harding's ``A Wasted Day'' or the SciFi story ``Captured Moments''. \textit{Newspapers} contain reports and longer reportages. \textit{Travel Guides} include both local histories and reflective pieces (e.g., ``Dublin and the Dubliners'').\footnote{See the full MASC corpus at: \url{https://anc.org/data/masc/corpus/browse-masc-data/}}

\begin{table}
    \centering
    \resizebox{\columnwidth}{!}{
    \begin{tabular}{l|crrrrrccc}
    \toprule
    \textbf{Dataset} &  \textbf{Period} & \textbf{N annotations} & \textbf{N words} & \textbf{$\bar{x}$ words/sentence} & \textbf{N annotators} & \multicolumn{3}{c}{\textbf{Krippendorff's $\alpha$}} \\
    \cmidrule(lr){7-9}
     & & & & & & \textbf{V} & \textbf{A} & \textbf{D} \\ 
    \midrule
    \textit{$\rightarrow$ Facebook} & 2012-2013 & 2,895 & 46,868 & 16.19 & 2 & .72 & .82 & - \\
    \midrule
       \textit{$\downarrow$ EmoBank} & 1990-2008 &  10,062 & 151,259 & 15.03 & 10  & .34 & .25 & .22 \\ \midrule
        \hspace{1.5em}Letters & & 1,413 & 21,639 & 15.31  & 10  & .35 & .25 & .25 \\
        \hspace{1.5em}Blog & & 1,336 &  20,874 & 15.62 & 10  & .32 & .22 & .18 \\
        \hspace{1.5em}Newspaper & & 1,314 & 25,992 & 19.78 & 10  & .30 & .22 & .22 \\
        \hspace{1.5em}Essays & & 1,135 & 26,349 & 23.21 & 10  & .33 & .21 & .21 \\
        \hspace{1.5em}Fiction & & 2,753 & 31,491 & 11.44 & 10  & .35 & .22 & .22 \\
        \hspace{1.5em}Travel-guides & & 919 & 17,154 & 18.67 & 10 & .28 & .23 & .23 \\
        \hspace{1.5em}SemEval & & 1,192 & 7,760 & 6.51 & 10 & .37 & .20 & .20 \\
        \midrule
       \textit{$\downarrow$ Fiction4} & 1798-1965 & 6,300 & 73,250 & 11.6 & $>=$2  & .67 & - & - \\ \midrule
        \hspace{1.5em}\worldflag[length=.28cm, width=.2cm]{DK} Hymns & 1798-1873 & 2,026 & 12,798 & 6.3 & 2 & .72 & - & - \\
        \hspace{1.5em}\worldflag[length=.28cm, width=.2cm]{DK} Fairy tales & 1837-1847 & 772 & 18,597 & 24.1 & 3 & .69 & - & - \\
        \hspace{1.5em}Prose & 1952 & 1,923 & 30,279 & 15.7 & 2  & .63 & - & - \\
        \hspace{1.5em}Poetry & 1965 & 1,579 & 11,576 & 7.3 & 3  & .59 & - & - \\        
        \bottomrule
    \end{tabular}}
        \caption{Datasets with valence annotation. Valence was annotated on a sentence basis, so `N annotations' indicates the number of sentences. `N annotators' indicates the number of annotators reported per sentence. IRR per dataset and category is shown in $\alpha$. Since \textit{EmoBank} lacks unique annotator IDs, we cannot correlate individual annotators' scores. Therefore, we use Krippendorff’s $\alpha$ measures agreement across V-A-D ratings per item in the full dataset and in subcategories. Only Emobank includes the full V-A-D annotation. Note that texts not indicated as Danish (flag) are all in English.}
    \label{tab:dataset_stats}
\end{table}

\FloatBarrier
\section{Train to test dataset correlations}
\label{sec:train-test-viz}
Visualizations of portability between datasets for valence. This figure is a visualization of \autoref{tab:subcategory-correlations}.

\FloatBarrier
\section{Beyond valence, visualized correlations}
\label{sec:arousal_dominance_vizes}

% \begin{figure}
%     \centering
%     \includegraphics[width=1\linewidth]{latex/arousal_scatter_matrix.pdf}
%     \caption{Relation between Concept Vector Projection scores (y-axis) and human scores of standardized arousal in the Emobank and Facebook corpora.}
%     \label{fig:placeholder}
% \end{figure}

% \begin{figure}
%     \centering
%     \includegraphics[width=0.5\linewidth]{latex/dominance_scatter.pdf}
%     \caption{Relation between Concept Vector Projection scores (y-axis) and human scores of standardized dominance in the Emobank corpus.}
%     \label{fig:placeholder}
% \end{figure}

\begin{figure}[htbp]
    \centering
    % Top: arousal
    \includegraphics[width=0.8\linewidth]{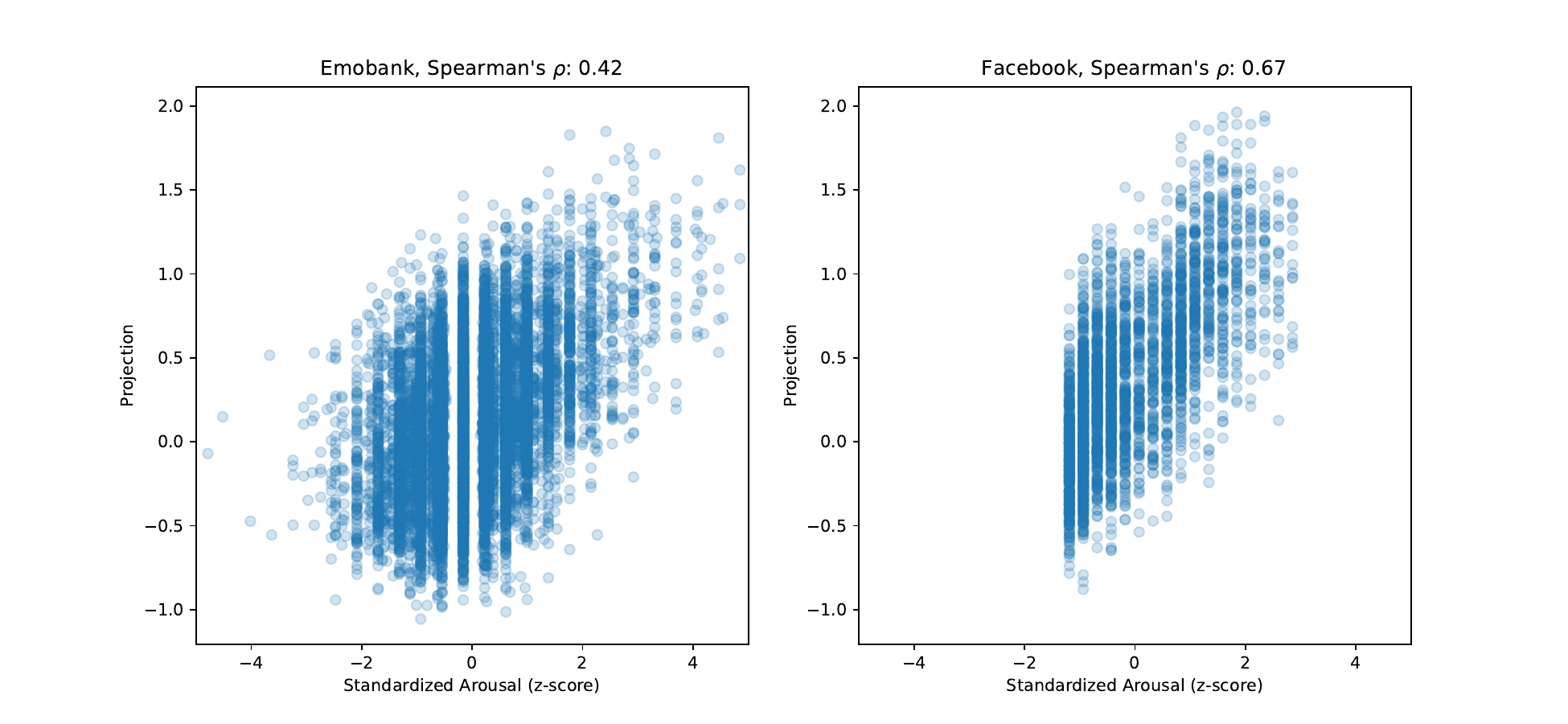}
    \vspace{0.5em} % optional spacing
    % Bottom: dominance
    \includegraphics[width=0.38\linewidth]{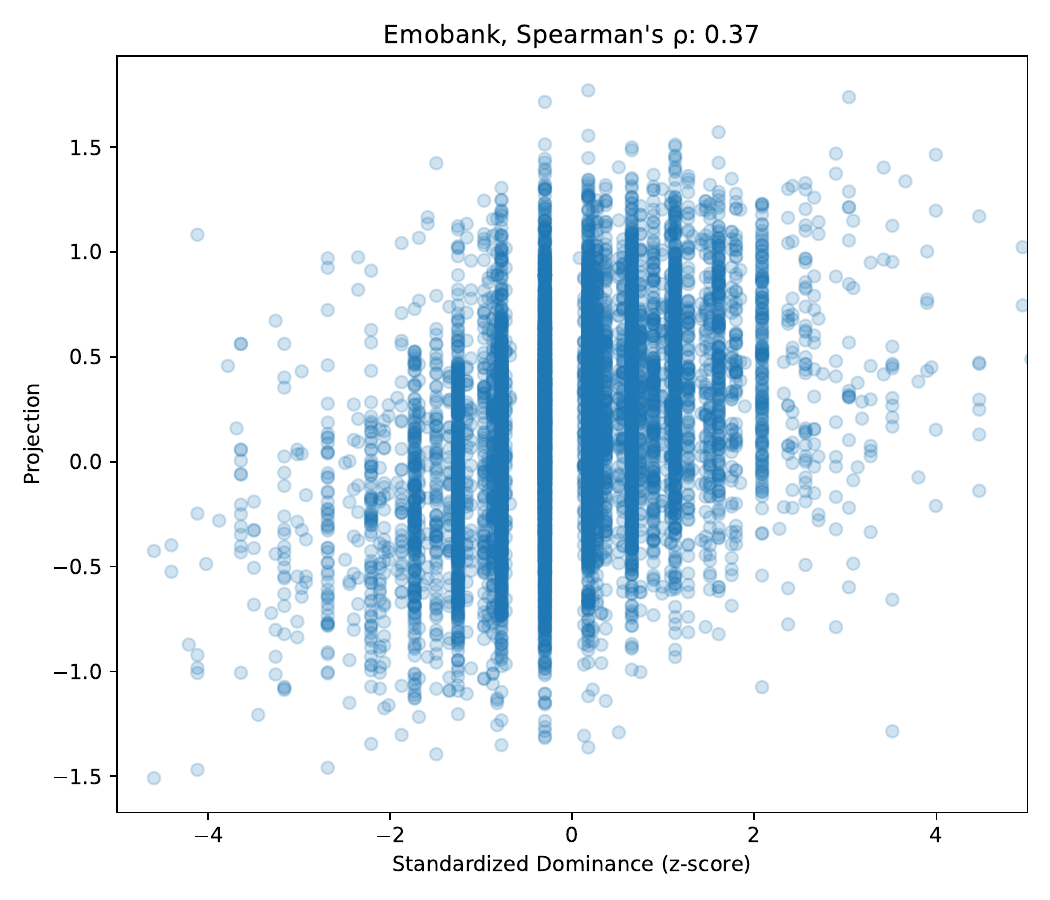}
    
    \caption{Top: Relation between Concept Vector Projection scores (y-axis) and human scores of standardized \textbf{arousal} in the Emobank and Facebook corpora. 
    Bottom: Relation between Concept Vector Projection scores (y-axis) and human scores of standardized \textbf{dominance} in the Emobank corpus.}
    \label{fig:arousal_dominance_visualized}
\end{figure}

\FloatBarrier
\section{Portability of Arousal}
Visualizations of portability between datasets for arousal. This figure is a visualization of \autoref{tab:subcategory-correlations}.
\label{app:portability-arousal}

\begin{figure}
    \centering
    \includegraphics[width=0.9\linewidth]{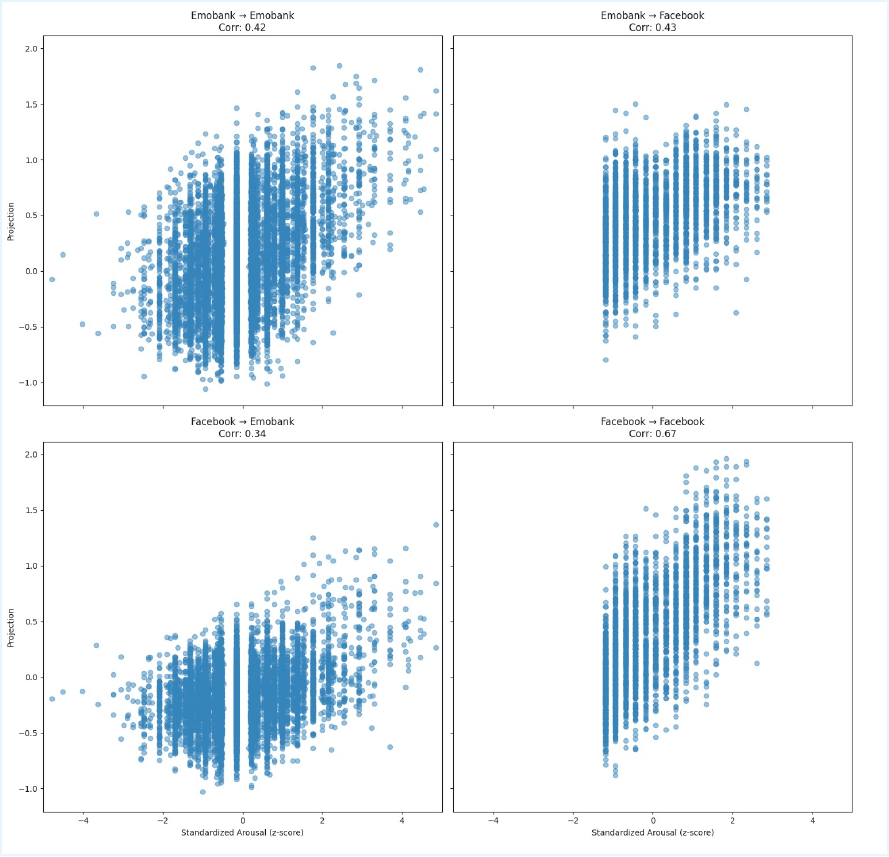}
    \caption{Relation between Concept Vector Projection scores (y-axis) and human scores (x-axis) on standardized arousal across datasets. A title such as Emobank$\rightarrow$ Facebook should be read as: Correlation between projections of arousal and human arousal ratings, when arousal vector is defined by the Emobank corpus and predictions are tested on the Facebook corpus.
    }
    \label{fig:scatters_all}
\end{figure}

\FloatBarrier
\section{Downstream differences between Human Annotators and Projection models}
\label{sec:Valence_Arousal_Correlation}
As a sanity check on downstream analysis using projection scores instead of human annotators we tested a simple hypothesis. That both high and low valence scores correlate with high arousal. This would imply that arousal only correlates with valence, when we use the absolute value of valence (i.e. distance from the mean). While the  slope of our linear regression varies between the two methods, we reach the same conclusion with both models. That there is a positive relation between absolute valence and arousal scores.

\begin{figure}
    \centering
    \includegraphics[width=0.9\linewidth]{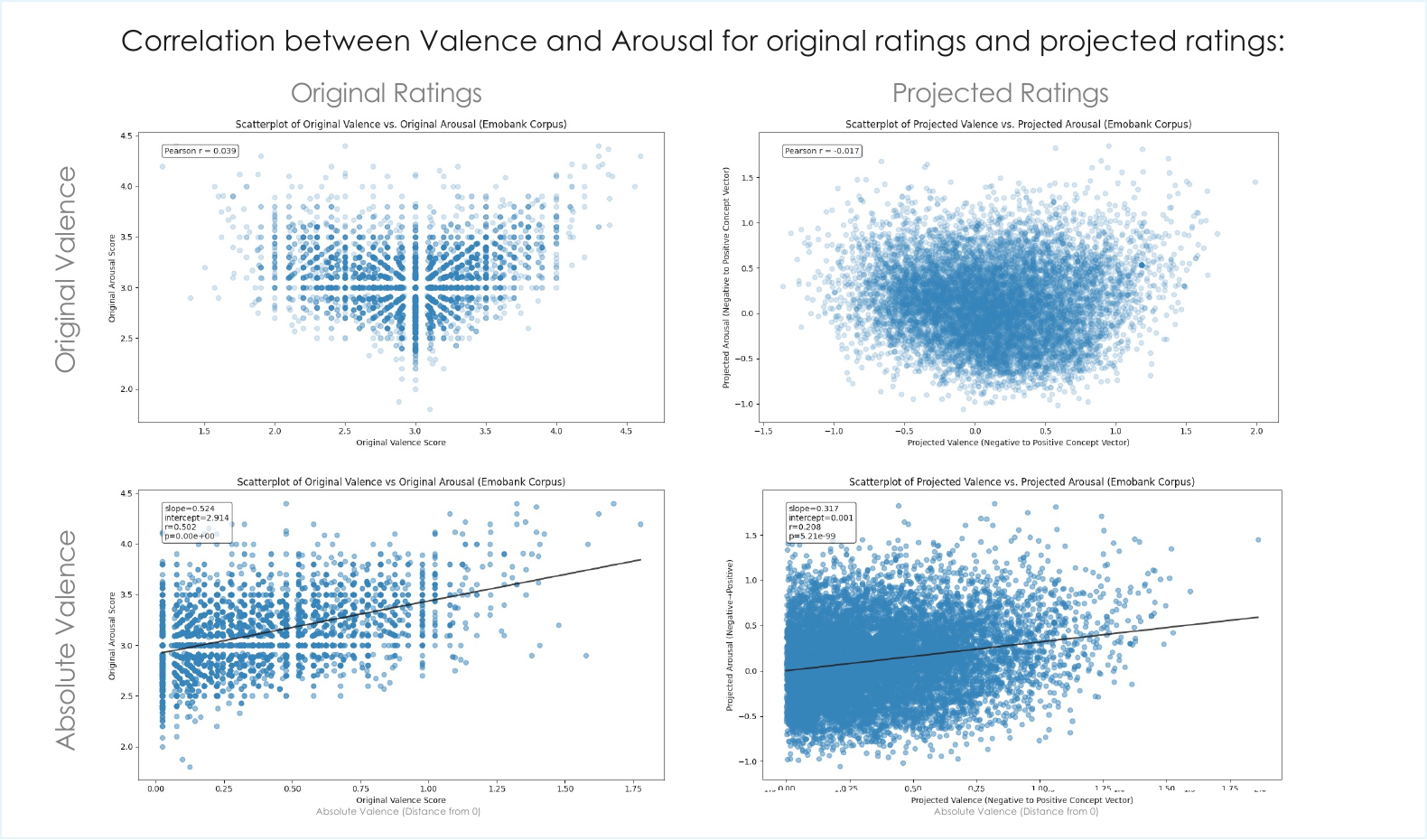}
    \caption{Scatterplots showing the relation between arousal and valence. The two top plots show no correlation between valence and arousal. The two bottom plots use absolute valence instead of valence, and depicts a positive significant relationship between absolute valence and arousal. Left side plots uses human annotations of EmoBank. Right side plots use projected ratings of EmoBank, and using the pos-neg vector defined on EmoBank.
    }
    \label{fig:scatters_arousal_valence}
\end{figure}

\end{document}